\documentclass[10pt,twocolumn,letterpaper]{article}

 \usepackage[pagenumbers]{style}

\usepackage[dvipsnames,table,xcdraw]{xcolor}

\usepackage{bm}
\usepackage{tabularray}
\usepackage{pifont}
\usepackage{stfloats}
\usepackage{float}
\usepackage{caption}
\usepackage{amsmath}
\usepackage{multirow}

\definecolor{cvprblue}{rgb}{0.21,0.49,0.74}
\usepackage[pagebackref,breaklinks,colorlinks,citecolor=cvprblue]{hyperref}

\title{COTR: Compact Occupancy TRansformer for Vision-based 3D Occupancy Prediction}

\author{Qihang Ma$^{1,*}$, Xin Tan$^{1,2,*}$, Yanyun Qu$^{3}$, Lizhuang Ma$^{1}$, Zhizhong Zhang$^{1,\dag}$, Yuan Xie$^{1,2}$ \\
$^1$ East China Normal University, Shanghai, China \\
$^2$ Chongqing Institute of East China Normal University, Chongqing, China \\
$^3$ Xiamen University, Fujian, China \\
{\tt qhma@stu.ecnu.edu.cn, \{xtan,zzzhang,yxie\}@cs.ecnu.edu.cn}, \\
{\tt ma-lz@cs.sjtu.edu.cn, yyqu@xmu.edu.cn}
}

\begin{document}

\maketitle
\renewcommand{\thefootnote}{}
\footnotetext{$^*$ Equal contribution, $^{\dag}$ Corresponding author.}

\begin{abstract}
The autonomous driving community has shown significant interest in 3D occupancy prediction, driven by its exceptional geometric perception and general object recognition capabilities. To achieve this, current works try to construct a Tri-Perspective View (TPV) or Occupancy (OCC) representation extending from the Bird-Eye-View perception. However, compressed views like TPV representation lose 3D geometry information while raw and sparse OCC representation requires heavy but redundant computational costs. To address the above limitations, we propose Compact Occupancy TRansformer (COTR), with a geometry-aware occupancy encoder and a semantic-aware group decoder to reconstruct a compact 3D OCC representation. The occupancy encoder first generates a compact geometrical OCC feature through efficient explicit-implicit view transformation. Then, the occupancy decoder further enhances the semantic discriminability of the compact OCC representation by a coarse-to-fine semantic grouping strategy. Empirical experiments show that there are evident performance gains across multiple baselines, \eg, COTR outperforms baselines with a relative improvement of 8\%-15\%, demonstrating the superiority of our method. The code is available at \textcolor{magenta}{https://github.com/NotACracker/COTR}.
\end{abstract}
    
\section{Introduction}
\label{sec:intro}
Vision-based 3D Occupancy Prediction aims to estimate the occupancy state of 3D voxels surrounding the ego-vehicle which provides a comprehensive 3D scene understanding~\cite{tong2023scene, tian2023occ3d, huang2023tri, wei2023surroundocc, Zhang_2023_ICCV}.
By dividing the whole space into voxels and predicting its occupancy and semantic information, the 3D occupancy network endows a universal object representation ability, where out-of-vocabulary objects and abnormity can be easily represented as \textit{[occupied; unknown]}.

\begin{figure}[t]
    \centering
    \setlength{\abovecaptionskip}{0pt}
    \includegraphics[width=\linewidth]{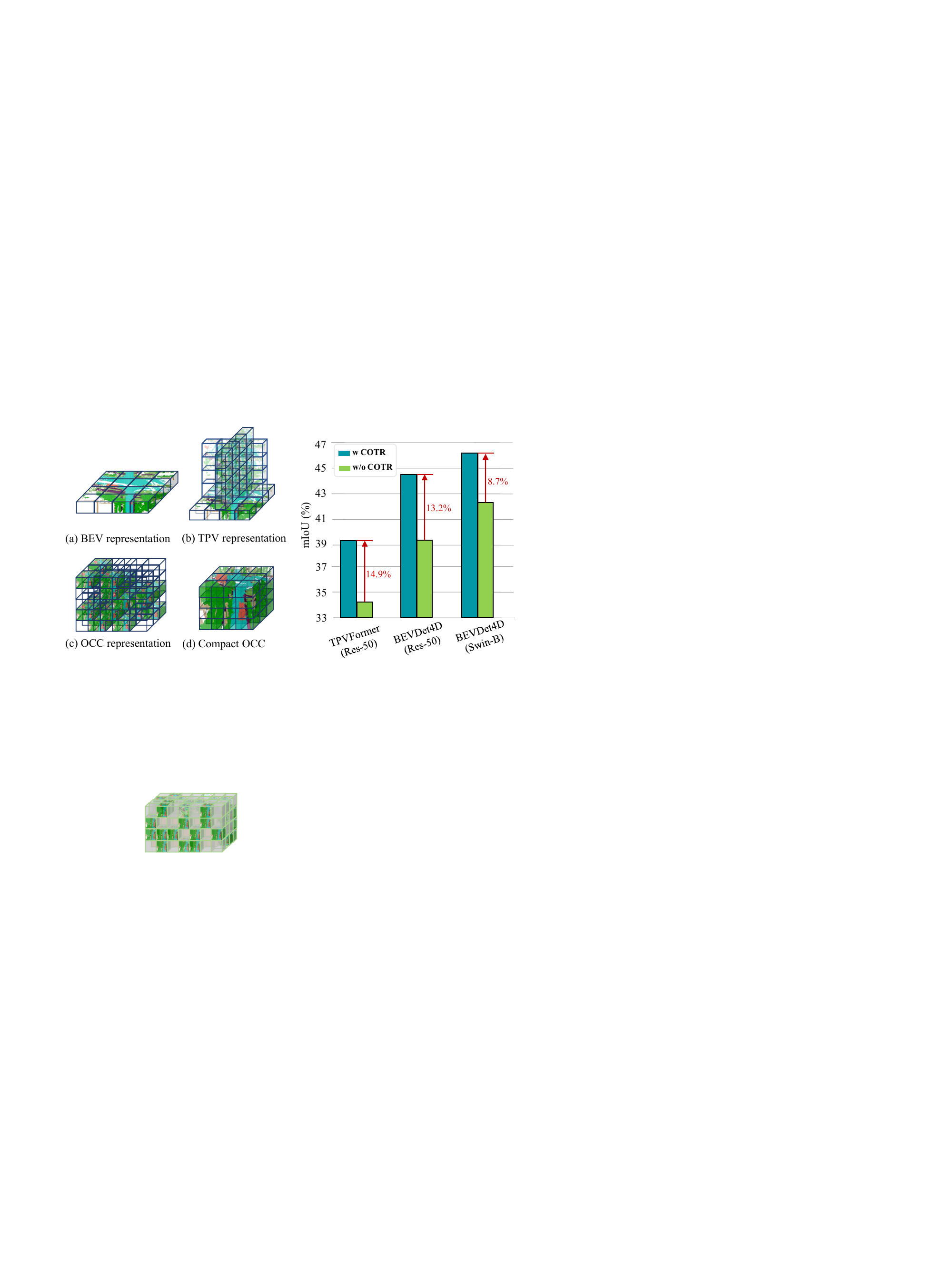}
    \caption{\textbf{Left:} Different representation for 3D perception. \textbf{Right:} The 3D Occupancy prediction results of different baselines with COTR on nuScenes~\cite{tian2023occ3d}. COTR outperforms baselines with a relative improvement of 8\%-15\%, demonstrating the superiority of our method.}
    \label{fig:motivation}
    \vspace{-1.5em}
\end{figure}

3D vision perception is now transitioning from Bird-Eye-View (BEV) perception~\cite{lu2019monocular, li2022bevformer, philion2020lift, huang2021bevdet, wang2022detr3d, liu2022petr} to Occupancy (OCC) perception~\cite{huang2023tri, wei2023surroundocc, Zhang_2023_ICCV, miao2023occdepth, Wang2023opencccupancy}. The BEV perception excels in 3D object detection tasks due to their unified representation abilities for multi-camera inputs, where the obstruction problem is extremely alleviated in the BEV plane. However, its deficiency in collapsing the height dimensions poses a challenge in preserving the requisite geometric information for a holistic understanding of the 3D scene. To alleviate this issue, \cite{huang2023tri} proposes a Tri-Perspective View (TPV) representation to delineate 3D scenes. Unfortunately, this introduces a new problem in that the compression along the horizontal dimension leads to significant object overlap.

From previous empirical studies, it appears that the compression on a specific dimension of 3D representations would lose substantial 3D geometry information. Thus the idea of 3D OCC representation is quite intriguing. It is generated by dividing the 3D space into uniform grids and therefore mapping the 3D physical world into a 3D OCC representation. Obviously, this representation costs huge computational resources than previous BEV or TPV representations. Moreover, due to the sparsity, the information density of such uncompressed representation is low, with numerous regions corresponding to free space in the physical world, resulting in significant redundancy.

Another issue is that the current 3D OCC representation lacks semantic discriminability which impedes the network's ability to successfully recognize rare objects. This primarily stems from the problem of class imbalance within the dataset which is common in the field of autonomous driving. To substantiate this assertion, we conducted a straightforward proxy experiment. In particular, for the network's prediction, we maintained the occupancy prediction unchanged and substituted the semantic prediction for non-empty regions with corresponding ground-truth semantics. The experimental results indicate an improvement of about 95\%, especially for the rare classes.

In this paper, we propose \textbf{\underline{C}}ompact \textbf{\underline{O}}ccupancy \textbf{\underline{TR}}ansformer, termed as \textbf{COTR}, which aims to construct a compact 3D OCC representation. Our objective is to preserve rich geometric information and minimize computational costs while concurrently enhancing semantic discriminability.

In this framework, we propose to construct a compact geometry-aware 3D occupancy representation through efficient explicit-implicit view transformation. Specifically, after generating a sparse yet high-resolution 3D OCC feature by Explicit View Transformation (EVT), we downsampled it to a compact OCC representation, a size merely 1/16 of the original, without any performance drop. Taking the compact OCC feature as input, Implicit View Transformation (IVT) further enriches it through spatial cross-attention and self-attention. Then, the updated OCC feature is upsampled to the original resolution for the downstream module. To recover the geometric details lost during downsampling, we configure the downsampling and upsampling processes into a U-Net architecture. Through such an approach, we substantially mitigate the sparsity of the OCC feature while simultaneously retaining geometric information and reducing unnecessary computational overhead and training time introduced by IVT. 

Secondly, we introduce a coarse-to-fine semantic-aware group decoder. We first divide the ground-truth labels into several groups based on semantic granularity and sample count. Then, for each semantic group, we generate corresponding mask queries and train the network based on group-wise one-to-many assignment. The grouping strategy results in balanced supervision signals, significantly enhancing the capability to recognize different classes, leading to a compact geometry- and semantic-aware OCC representation.

Our contributions can be summarized as follows:
\begin{itemize}
    \item We propose a geometry-aware occupancy encoder to construct a compact occupancy representation through efficient explicit-implicit view transformation. We can handle the sparsity of the occupancy feature while preserving the geometry information and reducing computation costs.
    \item We proposed a novel semantic-aware group decoder that significantly enhances the semantic discriminability of the compact occupancy feature. This group strategy balances the supervision signals and alleviates the suppression from common objects to rare objects.
    \item Our method has been embedded into several prevailing backbones, and experiments on the Occ3D-nuScenes dataset show that our approach achieves state-of-the-art performance. What's more, our method outperforms the backbones with a relative improvement of 8\%-15\%, as illustrated in Fig.~\ref{fig:motivation}.
\end{itemize}
\section{Related Work}
\label{sec:related_work}
\subsection{Vision-based BEV Perception}
\label{subsec:vb_bev_percep}
Over the recent years, vision-based Bird-Eye-View (BEV) perception has undergone significant development~\cite{li2022delving}, emerging as a crucial component in the autonomous driving community due to its cost-effectiveness, stability, and versatility. By transforming 2D image features to a unified and comprehensive 3D BEV representation through view transformation, various tasks, including 3D object detection and map segmentation, have been consolidated within a unified framework. View transformation can be broadly categorized into two types: one relies on explicit depth estimation to form a pseudo point cloud and construct the 3D space~\cite{philion2020lift, reading2021categorical, huang2022bevdet4d, li2023bevdepth, liang2022bevfusion}, while the other pre-defines the BEV space and implicitly models the depth information through spatial cross-attention~\cite{wang2022detr3d, li2022bevformer, yang2023bevformer, jiang2023polarformer, wang2023frustumformer}, mapping image features into the corresponding 3D positions. Although BEV perception excels in 3D object detection, it still encounters challenges in handling corner cases in driving scenarios, including irregular obstacles and out-of-vocabulary objects. To alleviate the aforementioned challenges, the 3D occupancy prediction task was proposed.

\subsection{3D Occupancy Prediction}
\label{subsec:3d_occ_pred}
The 3D occupancy prediction task has garnered significant attention due to its enhanced geometry information and superior capabilities in generalized object recognition compared to 3D object detection. TPVFormer~\cite{huang2023tri} adopts the concept of BEV perception, dividing 3D space into three perspective views and utilizing sparse point cloud supervision for 3D occupancy prediction. SurroundOcc~\cite{wei2023surroundocc} generates geometric information by expanding the height-dimensional of the BEV feature into occupancy features and conducting spatial cross-attention directly on them. Additionally, they introduce a new pipeline for constructing occupancy ground truth. 
OccNet~\cite{tong2023scene} bridges the end-to-end framework from perception to planning by constructing a general occupancy embedding. FBOcc~\cite{li2023fb} proposed a forward-backward view transformation module based on the BEV feature to address the limitations of different view transformations. While the aforementioned methods have made initial strides in the occupancy prediction task, a majority of them have largely adhered to the BEV perception framework and straightforwardly transformed BEV features to OCC features for the final prediction. They do not consider the sparsity and lack of semantic discriminability in the raw OCC representation.

\subsection{Semantic Scene Completion}
\label{subsec:ssc}
The definition of 3D occupancy prediction shares the most resemblance with Semantic Scene Completion (SSC)~\cite{song2017semantic, roldao2020lmscnet, chen20203d, li2020anisotropic, yan2021sparse}. MonoScene~\cite{cao2022monoscene} first proposed a framework that inferred dense geometry and semantics from a single monocular 2D RGB image. Voxformer~\cite{li2023voxformer} draws on the idea of BEV perception and employs depth estimation to construct a two-stage framework, which mitigates the overhead linked to attention computation. Occformer~\cite{Zhang_2023_ICCV} proposed a dual-path transformer and adopted the concept of mask classification for occupancy prediction. However, the performance of Voxformer relies on the robustness of the depth estimation, whereas Occformer's utilization of various transformers significantly increases the number of parameters. In this paper, we introduce an efficient framework to boost the performance of occupancy prediction while maintaining a low computational cost. 
\section{Methodology}
\label{sec:method}
\begin{figure*}[ht]
    \centering
    \setlength{\abovecaptionskip}{0pt}
    \includegraphics[width=\linewidth]{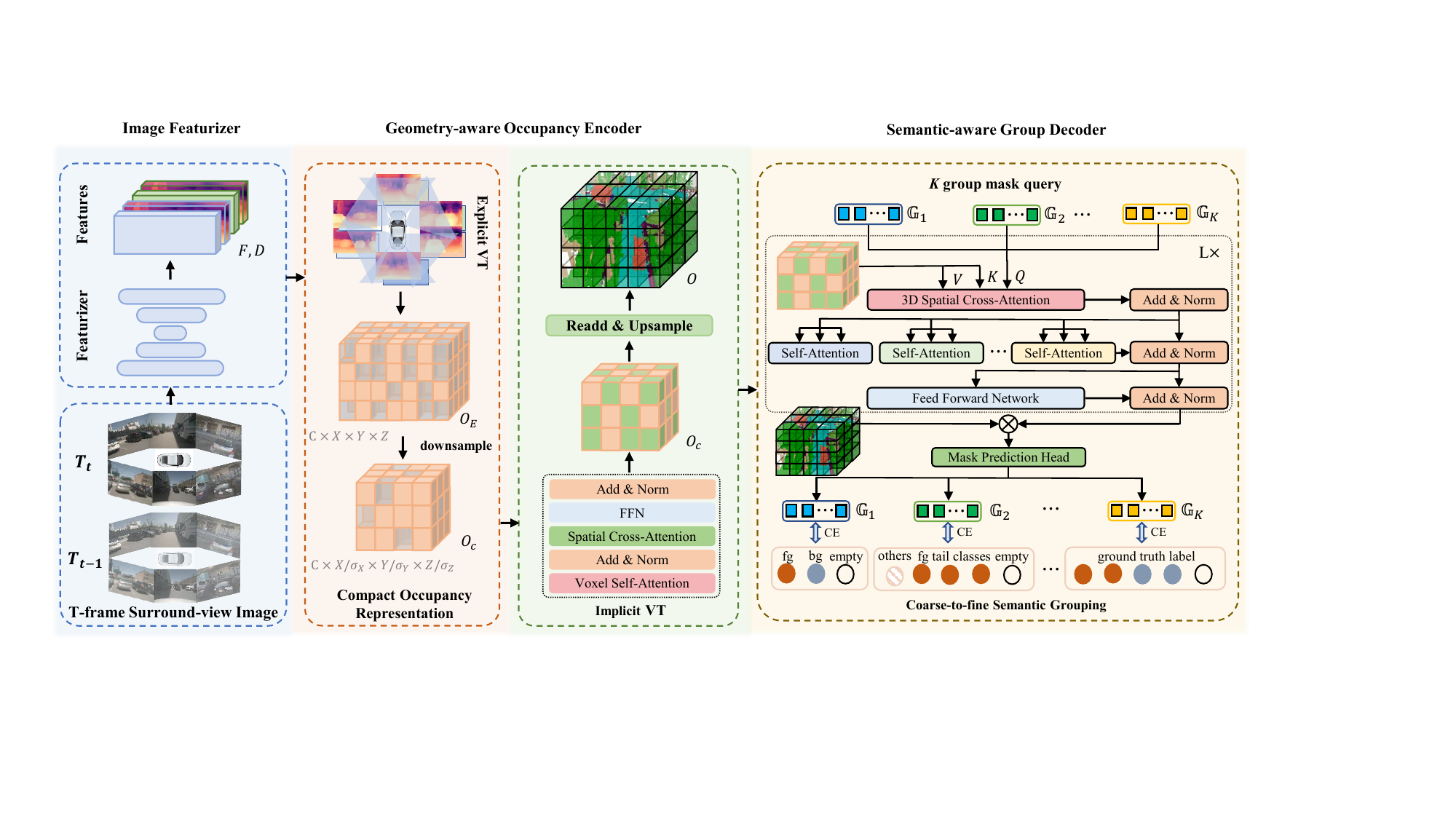}
    \caption{The overall architecture of COTR. $T$-frame surround-view images are first fed into the image featurizers to get the image features and depth distributions. Taking the image features and depth estimation as input, the geometry-aware occupancy encoder constructs a compact occupancy representation through efficient explicit-implicit view transformation. The semantic-aware group decoder utilizes a coarse-to-fine semantic grouping strategy cooperating with the Transformer-based mask classification to strongly strengthen the semantic discriminability of the compact occupancy representation.}
    \label{fig:overview}
    \vspace{-1.5em}
\end{figure*}
\label{subsec:overall}
\subsection{Preliminary}
\label{subsec:preliminary}

Given a sequence of multi-view image inputs, the goal of vision-centric 3D occupancy prediction is to estimate the state of 3D voxels surrounding the ego-vehicle. Specifically, the input of the task is a $T$-frame consequent sequence of images $\{I_{i,t} \in \mathbb{R}^{H_i \times W_i \times 3}\}$ from $N_c$ surround-view cameras, where $i\in\{1,\ldots,N_c\}$ and $t\in\{1,\ldots,T\}$. Besides, the camera intrinsic parameters $\{K_i\}$ and extrinsic parameters $\{[R_i|t_i]\}$ are also known for coordinate system conversions and ego-motion.

3D occupancy prediction aims to infer the states of each voxel, including \textit{occupancy} (\textit{[occupied]} or \textit{[empty]}) and \textit{semantics} (\textit{[category]} or \textit{[unknown]}) information. For example, a voxel on a car is annotated as \textit{[occupied; car]}, and a voxel in the free space is annotated as \textit{[empty; None]}. One primary advantage of the 3D occupancy prediction is to provide a universal object representation, where out-of-vocabulary objects and abnormity can be easily represented as \textit{[occupied; unknown]}.

\subsection{Overall Architecture}
\label{subsec:overall_arch}
An overview of the Compact Occupancy TRansforemr (COTR) is presented in Fig.~\ref{fig:overview}. The COTR mainly consists of three key modules: an image featurizer to extract image features and depth distributions, a geometry-aware occupancy encoder (Sec.~\ref{subsec:geo_occ_encoder}) to generate a compact occupancy representation through efficient explicit-implicit view transformation, and a semantic-aware group decoder (Sec.~\ref{subsec:sem_group_decoder}) to further enhance the semantic discriminability and geometry details of the compact OCC feature.

\textbf{Image Featurizer.}
The image featurizer aims to extract image features and depth distributions for multi-camera inputs, which provides the foundation of the geometry-aware occupancy encoder. Given a set of RGB images from multiple cameras, we first use a pretrained image backbone network (\eg, ResNet-50 \cite{he2016deep}) to extract image features $F=\{F_i \in \mathbb{R}^{C_F \times H \times W}\}_{i=1}^{N_c}$, where $F_i$ is the view features of $i$-th camera view and $N_c$ is the total number of cameras. Next, the depth distributions $D=\{D_i \in \mathbb{R}^{D_{bin} \times H \times W}\}_{i=1}^{N_c}$ can be obtained by feeding these image features $F$ into depth net.

\subsection{Geometry-aware Occupancy Encoder}
\label{subsec:geo_occ_encoder}
A key insight behind the occupancy task is that it could capture the fine-grained details of critical obstacles in the scene, such as the geometric structure of an object. To do so, we decided to use both explicit and implicit view transformation to generate a compact geometry-aware occupancy representation. In this section, we will first briefly review the explicit-implicit view transformation, and then intricately delineate how to construct the compact occupancy representation through efficient explicit-implicit fusion.

\textbf{Explicit-implicit View Transformation.}
Explicit and Implicit View Transformation is a crucial step in BEV perception~\cite{philion2020lift, huang2021bevdet, wang2022detr3d, li2022bevformer} to transform 2D image features to BEV representation. In order to construct a 3D representation that can preserve more 3D geometry information, we extend the Explicit-Implicit VT for OCC representation construction. Specifically, for EVT, the image features $F$ and depth distributions $D$ are computed by outer product $F \otimes D$ to obtain a pseudo point cloud feature $P \in \mathbb{R}^{N_cD_{bin}HW \times C}$. Then, instead of creating a BEV feature $B_{\rm E} \in \mathbb{R}^{C \times X \times Y}$, we directly generate a 3D OCC feature $O_{\rm E} \in \mathbb{R}^{C \times X \times Y \times Z}$ through voxel-pooling, where $(X, Y, Z)$ denotes the resolution of the 3D volume. For IVT, we predefine a group of grid-shaped learnable parameters $Q \in \mathbb{R}^{C \times X \times Y \times Z}$ as the queries of OCC, where each query is responsible for the corresponding grid cell in the 3D Occupancy space. Then the set of OCC queries will be updated through spatial cross-attention and self-attention to interact with the image features.

\textbf{Compact Occupancy Representation.}
With the adoption of EVT, we have already obtained a geometry-aware 3D OCC feature. A straightforward fusion approach is to directly incorporate this feature as the input to IVT. However, computing SCA with high-resolution 3D OCC features (\eg, $200 \times 200 \times 16$) incurs significant computational overhead. Additionally, due to the sparsity of 3D space, the computation in the majority of free space is also ineffective. Therefore, we downsampled the high-resolution but sparse OCC feature $O_{\rm E}$ to a compact OCC representation $O_c \in \mathbb{R}^{C \times \frac{X}{\sigma_x} \times \frac{Y}{\sigma_y} \times \frac{Z}{\sigma_z}}$, where $\sigma_x, \sigma_y, \sigma_z$ denote the downsample ratios. Taking each 3D voxel of the compact OCC $O_c$ as queries, IVT completes the sparse regions and further enriches the geometric details therein. Compared to a standard encoder which learns a set of queries from scratch, this operation significantly saves extra training time and reduces computational overhead. Then the compact $O_c$ is upsampled to original resolution $O \in \mathbb{R}^{C_{Q} \times X \times Y \times Z}$ for final occupancy prediction. Since the downsampling operation inevitably introduces information loss, especially for small objects, we constructed the downsampling and upsampling processes into a Unet~\cite{ronneberger2015u} architecture to mitigate this problem. In practice, we construct a compact OCC representation with a size merely 1/16 of the original while achieving a better performance.

\begin{figure*} [ht]
	\centering
	\setlength{\abovecaptionskip}{0pt}
    \includegraphics[width=\linewidth]{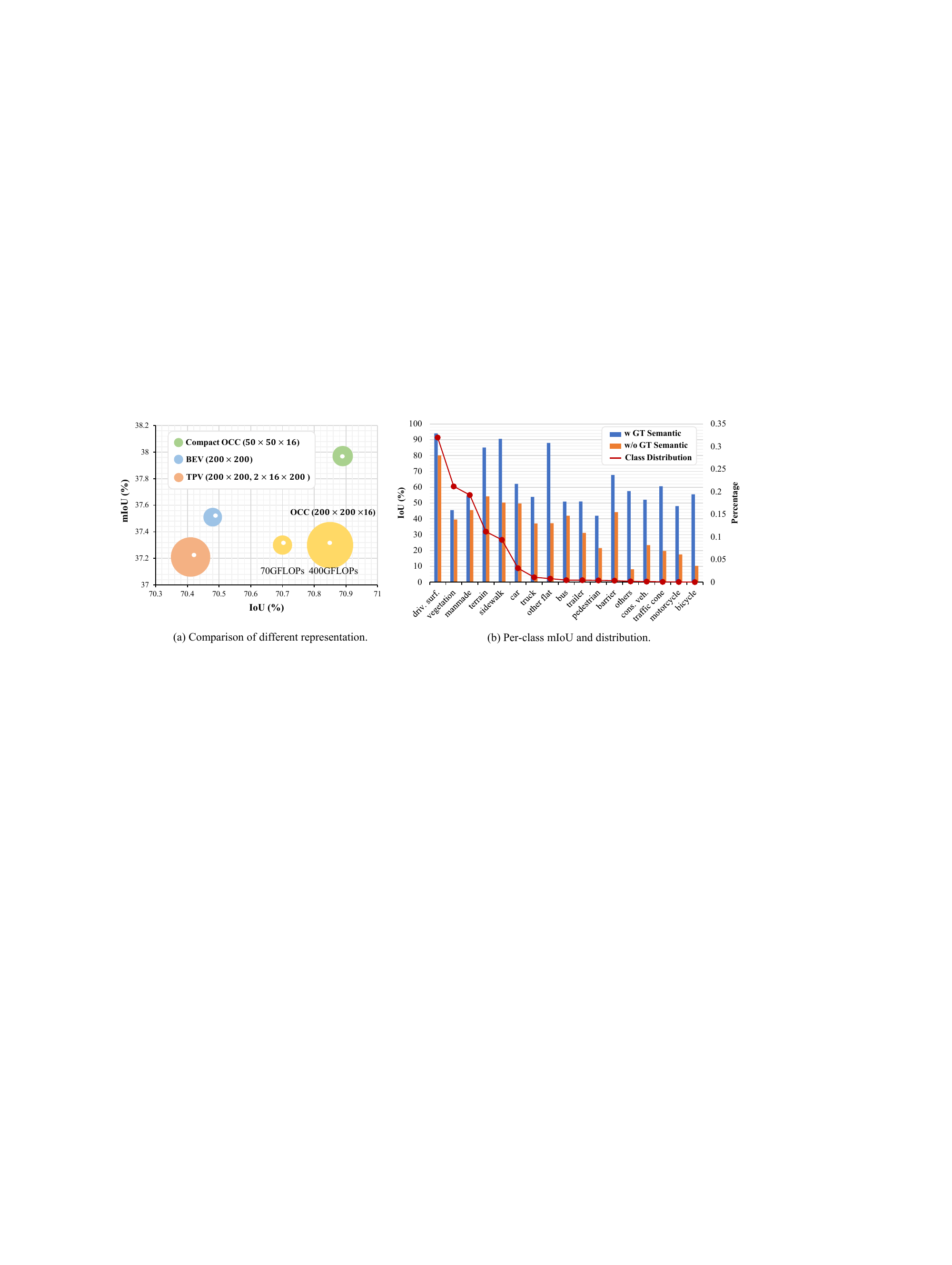}
	\caption{Proxy experiments. (a) depicts the comparison of different occupancy representations. The compact 3D OCC representation achieves a balance between performance and computational cost. (b) reports the per-class mIoU and distribution with and without using ground-truth semantic labels.}
	\label{fig:proxy_exp}
    \vspace{-1.5em}
\end{figure*}

\textbf{Discussion.}
There are three overarching advantages to employing the compact occupancy representation. Firstly, the 3D feature representation enjoys a natural geometric superiority over 2D BEV or TPV~\cite{huang2023tri}. As illustrated in Fig.~\ref{fig:proxy_exp}~(a), the compact OCC representation achieves the best IoU score. Secondly, the compact OCC representation effectively alleviates the sparsity inherent in high-resolution OCC features. For outdoor autonomous driving datasets such as Nuscenes~\cite{caesar2020nuscenes}, SemanticKITTI~\cite{song2017semantic}, and Waymo~\cite{sun2020scalability}, the proportion of free space stands at 78\%, 93\%, and 92\%, respectively. The compact OCC representation denotes a compressed spatial domain, enriched features, and expanded receptive fields. Thirdly, the computational overhead is significantly diminished. As depicted in Fig.~\ref{fig:proxy_exp}~(a), the computational cost of a raw high-resolution OCC representation is approximately 500\% higher than that of a compact OCC representation.

\subsection{Semantic-aware Group Decoder}
\label{subsec:sem_group_decoder}
In this section, we present our semantic-aware group decoder, which further enhances the geometric occupation of the compact OCC feature while greatly improving semantic discriminability. We will commence with a proxy experiment designed to substantiate our assertion that the occupancy feature lacks semantic discriminability, which significantly impedes the recognition of rare objects. Subsequently, we will delve into the details of our coarse-to-fine semantic grouping strategy.

\textbf{Proxy Experiment.} To demonstrate that the occupancy feature lacks semantic discriminability, we replace the semantic prediction in our occupancy prediction with the ground-truth label in the corresponding position. As shown in Fig.~\ref{fig:proxy_exp}~(b), the IoU scores are greatly improved, especially for the tail class. This drives us to look for a new approach to greatly enhance the semantic discrimination of the occupancy feature.

\textbf{Transformer Decoder.}
Inspired by MaskFormer~\cite{cheng2021per}, We convert the occupancy prediction to the form of mask classification. This form of prediction splits the occupancy prediction into two sub-problems, which is convenient for us to address the semantic obfuscation problem. To do so, we replace the image feature in the Transformer decoder with the compact occupancy feature $O_c$ from our geometry-aware occupancy encoder. In addition, we replace the original encoder-decoder global self-attention layer with 3D SCA to further reduce the computational cost. The 3D Spatial Cross Attention (3D-SCA) can be formulated as:
\begin{equation}
    {\rm 3D\mbox{-}SCA}(Q_m, O_c)=\sum_{i=1}^{\mathcal{N}_{\rm ref}}f(Q_m, \bold{p}_i, O_c),
\end{equation}
where $f(\cdot)$ is the deformable attention function, $Q_m \in \mathbb{R}^{N_q \times C_m}$ is $N_q$ learnable mask queries, $\bold{p}$ denotes the sample location in the 3D occupancy space and we sample $\mathcal{N}_{\rm ref}$ 3D occcupancy features for each mask query $Q_m$. The updated mask queries are then fed into self-attention to interact with other mask queries. At the end of each iteration, each mask query $q_i \in Q_m$ will be projected to predict its class probability $\{p_i \in \delta^{K+1}\}_{i=1}^{N_q}$ and the mask embedding $\mathcal{E}_m$. Then, the latter is further converted to a binary occupancy mask $m_i \in [0,1]^{X \times Y \times Z}$ through a dot product with the occupancy feature $O$ and a sigmoid function:
\begin{equation}
    m_i[x,y,z] = {\rm sigmoid}(\mathcal{E}_m[:,i]^{\rm T} \cdot O[:,x,y,z]).
\end{equation}

\textbf{Coarse-to-Fine Semantic Grouping.}
Because of the imbalanced data distribution, the classifier that predicts the class probability would make the classification scores for low-shot categories much smaller than those of many-shot categories, resulting in semantic misclassification. In order to enhance the supervision signal of the rare classes, we first adopt a group-wise one-to-many assignment according to ~\cite{chen2023group}, which aims to enable each mask query to obtain multiple positive matching pairs. However, experiments show that such a simple grouping strategy is ineffective and can not bring performance improvement. Inspired by~\cite{li2020overcoming, xiang2020learning, cai2021ace}, based on the group-wise one-to-many assignment, we further introduce a Coarse-to-Fine Semantic Grouping strategy. This involves partitioning mask queries into $K$ groups, with each group supervised by dividing semantic categories into $K$ ground truth (gt) label groups based on semantic granularity and sample count, aimed at balancing the supervision signal in each group.

Concretely, we first divided the categories into $\mathbb{G}_1 = $~\textit{\{``foreground", ``background", ``empty"\}}. Then for the foreground or background category, we again grouped the categories based on the number of training samples. We assign category $i$ into group $\mathbb{G}_n$ if:
\begin{equation}
    N_n^l < \mathcal{N}(i) \leq N_n^h,~~~~n\in[2,K-1],
\end{equation}
where $\mathcal{N}(i)$ is the number of training samples for category $i$, and $N_n^l$ and $N_n^h$ are hyper-parameters that determine minimal and maximal sample numbers for group $n$. 

In order to keep only one group during inference, we manually set the $\mathbb{G}_K$ to the original ground truth label and  $\mathbb{G}_1$ to \textit{\{``foreground", ``background", ``empty"\}}, thus there is no extra cost compared with single-group models. Besides, during training, in order for each class to be supervised in each group, we categorize the classes not assigned to the current group as \textit{``others"}. For example, if the foreground classes \textit{motorcycle} and \textit{bicycle} are assigned in the same group, the ground truth label for this group is \textit{\{``others", ``motorcycle", ``bicycle", ``empty" \}}. For label groups like $\mathbb{G}_1$, category categorization is coarse, while in the subsequent groups, it becomes more fine-grained.

\section{Experiments}
\label{sec:experiments}

\definecolor{nbarrier}{RGB}{255, 120, 50}
\definecolor{nbicycle}{RGB}{255, 192, 203}
\definecolor{nbus}{RGB}{255, 255, 0}
\definecolor{ncar}{RGB}{0, 150, 245}
\definecolor{nconstruct}{RGB}{0, 255, 255}
\definecolor{nmotor}{RGB}{200, 180, 0}
\definecolor{npedestrian}{RGB}{255, 0, 255}
\definecolor{ntraffic}{RGB}{255, 240, 150}
\definecolor{ntrailer}{RGB}{135, 60, 0}
\definecolor{ntruck}{RGB}{255, 0, 0}
\definecolor{ndriveable}{RGB}{213, 213, 213}
\definecolor{nother}{RGB}{139, 137, 137}
\definecolor{nsidewalk}{RGB}{75, 0, 75}
\definecolor{nterrain}{RGB}{150, 240, 80}
\definecolor{nmanmade}{RGB}{160, 32, 240}
\definecolor{nvegetation}{RGB}{0, 175, 0}
\definecolor{nothers}{RGB}{0, 0, 0}

\begin{table*}[ht]
	\footnotesize
    \setlength{\abovecaptionskip}{0pt}
	\setlength{\tabcolsep}{0.02\linewidth}
	\newcommand{\classfreq}[1]{{~\tiny(\nuscenesfreq{#1}\%)}}  %
    \begin{center}
	\begin{tabular}{l|c|c|c|c|c|cc}
		\toprule
		Method
		& Venue & Image Backbone & Image Size & Epoch & Visible Mask & IoU (\%) & mIoU (\%)
		\\
		\midrule
        MonoScene~\cite{cao2022monoscene} & CVPR'{\color{blue}22} & ResNet-101 & $928\times600$ & 24 & \ding{56} & - & 6.1  \\
        OccFormer*~\cite{Zhang_2023_ICCV} & ICCV'{\color{blue}23} & ResNet-50 & $256\times704$ & 24 & \ding{56} & 30.1 & 20.4  \\
        BEVFormer~\cite{li2022bevformer} & ECCV{\color{blue}22} & ResNet-101 & $928\times600$ & 24 & \ding{56} & - & 26.9  \\
        CTF-Occ~\cite{tian2023occ3d} & arXiv'{\color{blue}23} & ResNet-101 & $928\times600$ & 24 & \ding{56} & - & 28.5  \\
        VoxFormer~\cite{li2023voxformer} & CVPR'{\color{blue}23} & ResNet-101 & $900\times1600$ & 24 & \ding{52} & - & 40.7  \\
        SurroundOcc~\cite{wei2023surroundocc} & ICCV'{\color{blue}23} & InternImage-B & $900\times1600$ & 24 & \ding{52} & - & 40.7  \\
        FBOcc$\dagger$~\cite{li2023fb} & ICCV'{\color{blue}23} & ResNet-50 & $256\times704$ & 20 & \ding{52} & - & 42.1  \\ 
        \midrule
        TPVFormer*~\cite{huang2023tri} & CVPR'{\color{blue}23} & ResNet-50 & $900\times1600$ & 24 & \ding{52} & 66.8 & 34.2  \\
        \rowcolor[HTML]{EFEFEF}
        TPVFormer + COTR & - & ResNet-50 & $256\times704$ & 24 & \ding{52} & \textbf{70.6} & \textbf{39.3}  \\
        \midrule
        SurroundOcc*~\cite{wei2023surroundocc} & ICCV'{\color{blue}23} & ResNet-101 & $900\times1600$ & 24 & \ding{52} & 65.5 & 34.6 \\
        \rowcolor[HTML]{EFEFEF}
        SurroundOcc + COTR & - & ResNet-50 & $256\times704$ & 24 & \ding{52} & \textbf{71.0} & \textbf{39.3}\\
        \midrule
        OccFormer*~\cite{Zhang_2023_ICCV} & ICCV'{\color{blue}23} & ResNet-50 & $256\times704$ & 24 & \ding{52} & 70.1 & 37.4 \\
        \rowcolor[HTML]{EFEFEF}
        OccFormer + COTR & - & ResNet-50 & $256\times704$ & 24 & \ding{52} & \textbf{71.7} & \textbf{41.2}\\
        \midrule
        BEVDet4D$\ddagger$~\cite{huang2022bevdet4d} & arXiv'{\color{blue}22} & ResNet-50 & $384 \times 704$ & 24 & \ding{52} & 73.8 & 39.3 \\
        \rowcolor[HTML]{EFEFEF}
        BEVDet4D + COTR & - & ResNet-50 & $256\times704$ & 24 & \ding{52} & \textbf{75.0} & \textbf{44.5} \\
        \midrule
        BEVDet4D$\ddagger$~\cite{huang2022bevdet4d} & arXiv'{\color{blue}22} & SwinTransformer-B & $512 \times 1408$ & 36 & \ding{52} & 72.3 & 42.5 \\
        \rowcolor[HTML]{EFEFEF}
        BEVDet4D + COTR & - & SwinTransformer-B & $512 \times 1408$ & 24 & \ding{52} & \textbf{74.9} & \textbf{46.2} \\
		\bottomrule
	\end{tabular}
    \end{center}
    \caption{\textbf{3D Occupancy prediction performance on the Occ3D-nuScenes dataset.} $\dagger$ with test-time augmentation. $\ddagger$ means the performance is reported by its official code. * means the performance is achieved by our implementation using its official code. Visible Mask means whether models are trained with visible masks.}
	\label{tab:nuscene_sota}
    \vspace{-1.5em}
\end{table*}

\subsection{Experimental Setup}
\label{subsec:experimental_setup}

\textbf{Dataset.} Occ3D-nuScenes~\cite{tian2023occ3d} is a large-scale autonomous driving dataset, which contains 700 training scenes and 150 validation scenes. Each frame contains a 32-beam LiDAR point cloud and six RGB images captured by six cameras from different views of LiDAR with dense voxel-wise semantic occupancy annotations. The occupancy scope is defined as -40$m$ to 40$m$ for the X and Y axis, and -1$m$ to 5.4$m$ for the Z axis in the ego coordinate. The voxel size is $0.4m \times 0.4m \times 0.4m$ for the occupancy label. The semantic labels contain 17 categories, including 16 known object classes with an additional ``empty'' class. 

\textbf{Implementation Details.} By adhering to common practices~\cite{huang2021bevdet,li2023bevstereo,li2023fb}, we default to using ResNet-50~\cite{he2016deep} as the image backbone, and the image size is resized to ($256 \times 704$) for Occ3D-nuScenes. For explicit view transformation, we adopt BEVStereo~\cite{li2023bevstereo} which depth estimation is supervised from sparse LiDAR. The resolution of the occupancy feature from explicit view transformation is $200 \times 200 \times 16$ with a feature dimension $C$ of 32, and the downsample ratios are $\sigma_x=\sigma_y=4,\sigma_z=1$ with an embedding dimension $C_Q$ of 256. We use 8 attention heads for both self- and cross-attention and set $\mathcal{N}_{ref}=4$ for both 2D and 3D SCA. We simply generate $K=6$ group gt labels and mask queries for the Group Occupancy Decoder, which We simply divided both the foreground and background classes into two separate groups, resulting in a total of 4 groups. This division was based on the median number of training samples.

We also integrate our method into four main-stream occupancy models BEVDet4D~\cite{huang2022bevdet4d}, TPVFormer~\cite{huang2023tri}, SurroundOcc~\cite{wei2023surroundocc} and OccFormer~\cite{Zhang_2023_ICCV} to demonstrate the effectiveness of our approach. Unless otherwise specified, all models are trained for 24 epochs using AdamW optimizer~\cite{loshchilov2017decoupled}, in which gradient clip is exploited with learning rate 2e-4.

\begin{figure*}[ht]
    \centering
    \includegraphics[width=\linewidth]{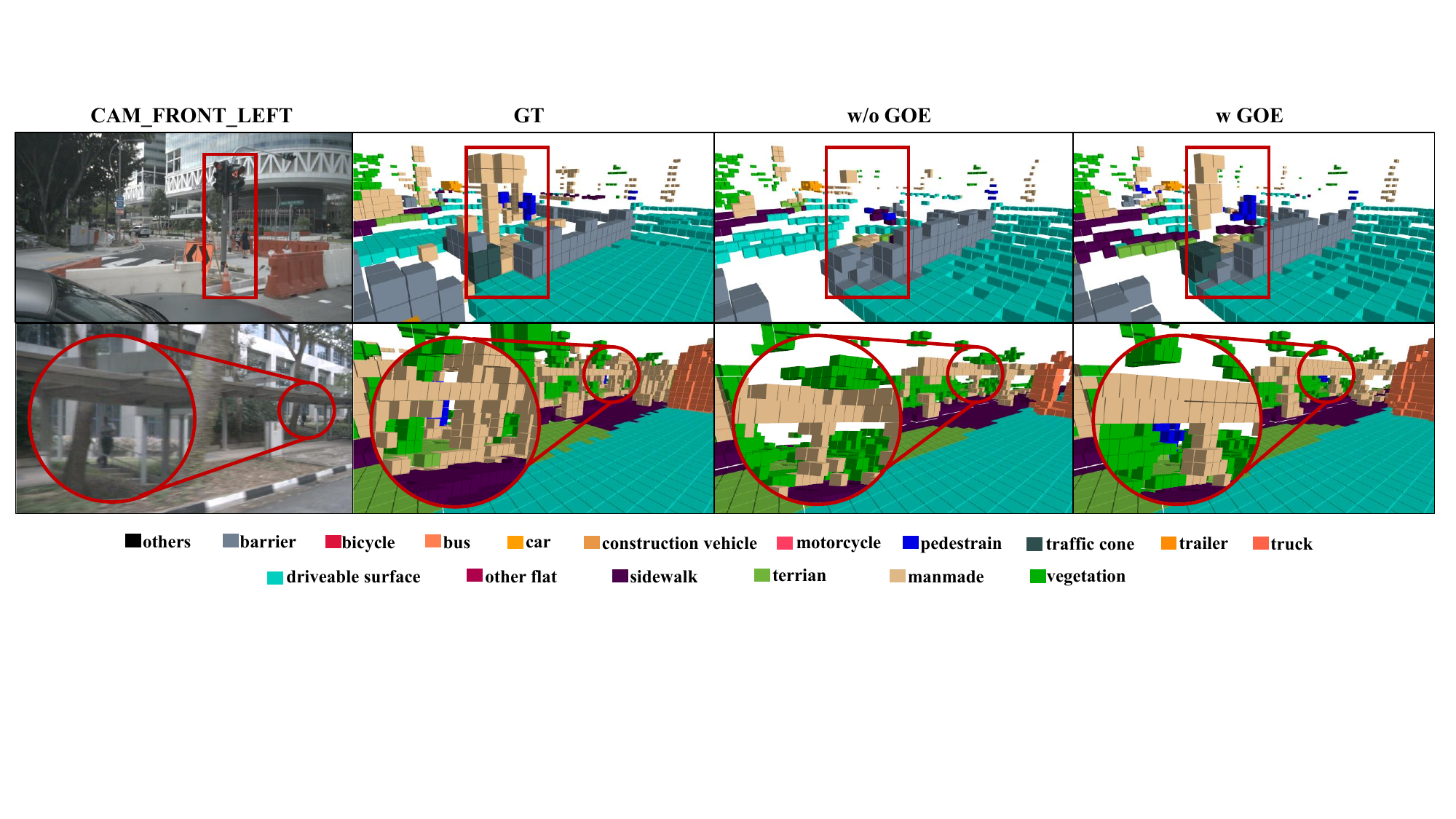}
    \setlength{\abovecaptionskip}{-1.0em}
    \caption{Qualitative results comparison between baseline and our Geometry-aware Occupancy Encoder. The results demonstrate that the compact occupancy representation is able to capture more precise geometrical details for slimmer objects (\eg, pedestrians and poles) and is robust to occlusions by combining the advantages of implicit and explicit view transformation.}
    \label{fig:EI_ablation}
    \vspace{-0.5em}
\end{figure*}
\begin{table}[]
\begin{center}
\begin{tabular}{ccc|cccc}
\toprule
\multicolumn{3}{c|}{Component} & \multicolumn{4}{c}{Metric} \\
\midrule
GOE & TD & CFSG & \multicolumn{2}{c}{IoU (\%)} & \multicolumn{2}{c}{mIoU (\%)} \\
\midrule
 &  &  & 70.36 & - & 36.01 & -\\
\ding{52} &  &  & 70.89 & {\color{blue}+0.53} & 37.97 & {\color{blue}+1.98}\\
 & \ding{52} &  & 69.76 & {\color{blue}-0.60} & 38.43 & {\color{blue}+2.42} \\
\ding{52} & \ding{52} &  & 71.74 & {\color{blue}+1.38} & 40.22 & {\color{blue}+4.21}\\
\rowcolor[HTML]{EFEFEF}
\ding{52} & \ding{52} & \ding{52} & \textbf{72.08} & {\color{blue}\textbf{+1.72}} & \textbf{41.39} & {\color{blue}\textbf{+5.38}}\\
\bottomrule
\end{tabular}
\end{center}
\setlength{\abovecaptionskip}{0pt}
\caption{\textbf{Ablation study on the each component.} GOE denotes the Geometry-aware Occupancy Encoder, TD denotes the Transformer Decoder and CFSG means Course-to-Fine Semantic Grouping. All models are trained without long-term temporal information.}
\label{tab:ablation_compo}
\vspace{-1.75em}
\end{table}
\subsection{Comparing with SOTA methods}
\label{subsec:compare_sota}

\textbf{Occ3D-nuScenes.} As shown in Table~\ref{tab:nuscene_sota}, we report the quantitative comparison of existing state-of-the-art methods for 3D occupancy prediction tasks on Occ3D-nuScenes. We integrate our method into TPVFormer, SurroundOcc, OccFormer and BEVDet4D, and our approach yields significant performance improvements in both geometric completion and semantic segmentation, surpassing the baseline by 3.8\%, 5.5\%, 1.6\%, 1.2\% in IoU and 5.1\%, 4.7\%, 3.8\%, 5.2\% in mIoU. Notably, our approach based on BEVDet4D utilizes a smaller backbone (ResNet-50) and smaller image input size ($256 \times 704$) to achieve mIoU scores of 44.5\%, which outperforms both Voxformer~\cite{li2023voxformer} (ResNet-101, $900 \times 1600$) and SurroundOcc~\cite{wei2023surroundocc} (InternImage-B) by 3.8\%. This demonstrates that our method mines more information with a small number of parameters by designing components specifically for 3D occupancy prediction tasks. Compared to the state-of-the-art FBOcc~\cite{li2023fb}, BEVDet4D with COTR surpasses it by 2.3\% even without test-time augmentation. In addition, we also scale up the image backbone to SwinTransformer-B~\cite{liu2021swin} and Image Size to $512 \times 1408$. The experimental results demonstrate that our method consistently brings performance improvements, even with larger model sizes.

\begin{table}[]
\resizebox{\linewidth}{!}{
\begin{tabular}{c|c|cccc}
\toprule
Groups          & LT & \multicolumn{2}{c}{IoU (\%)} & \multicolumn{2}{c}{mIoU (\%)} \\
\midrule
\{\textit{gt label}\}  & \ding{56} & 71.74 & - & 40.22 & -          \\
$10 \times$ \{\textit{gt label}\} &  \ding{56} & 71.47 & {\color{blue}-0.27} & 40.17 & {\color{blue}-0.05}      \\
\{\textit{fg}, \textit{bg}, \textit{empty}\}, \{\textit{gt label}\}  & \ding{56} & 72.11 & {\color{blue}+0.37} & 40.61 & {\color{blue}+0.39} \\
\rowcolor[HTML]{EFEFEF}
\{\textit{fg}, \textit{bg}, \textit{empty}\}, $\ldots$, \{\textit{gt label}\} & \ding{56} & 72.08 & {\color{blue}+0.34} & 41.39  & {\color{blue}+1.17}      \\
\midrule
\{\textit{gt label}\} & \ding{52} & 73.68 & - & 42.70 & -          \\
\rowcolor[HTML]{EFEFEF}
\{\textit{fg}, \textit{bg}, \textit{empty}\}, $\ldots$, \{\textit{gt label}\} & \ding{52} & 75.01 & {\color{blue}+1.33} & 44.45 & {\color{blue}+1.75}       \\
\bottomrule
\end{tabular}
}
\setlength{\abovecaptionskip}{5pt}
\caption{\textbf{Ablation study on the number of Semantic Groups.} $10 \times$ \{\textit{gt label}\} means we replicated the original ground truth labels ten times to verify that the performance improvement was due to the increase in model parameters. LT means models are trained with long-term temporal information.}
\label{tab:ablation_group_num}
\vspace{-1.5em}
\end{table}
\begin{figure}[]
    \centering
    \includegraphics[width=\linewidth]{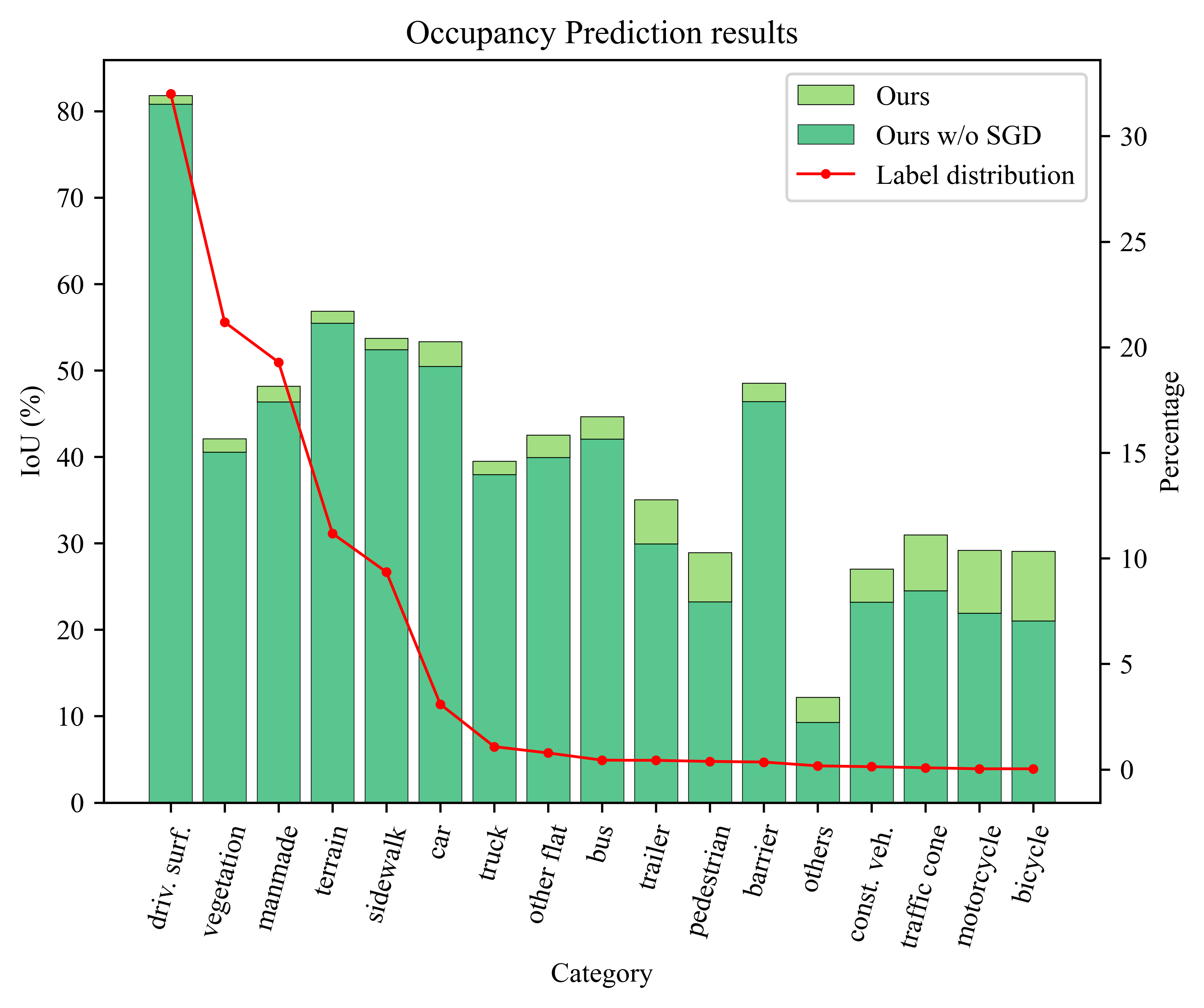}
    \setlength{\abovecaptionskip}{-1.5em}
    \caption{The occupancy prediction results with label distribution. It is clear to see that the Semantic-aware Group Decoder (SGD) gives a big performance boost to the rare class.}
    \label{fig:god_longtail_cls}
    \vspace{-2.5em}
\end{figure}
\subsection{Ablation study}
To delve into the effect of different modules, we conduct ablation experiments on Occ3d-nuScenes~\cite{tian2023occ3d} based on BEVDet4D~\cite{huang2022bevdet4d}.
\begin{figure*}[]
    \centering
    \includegraphics[width=\linewidth]{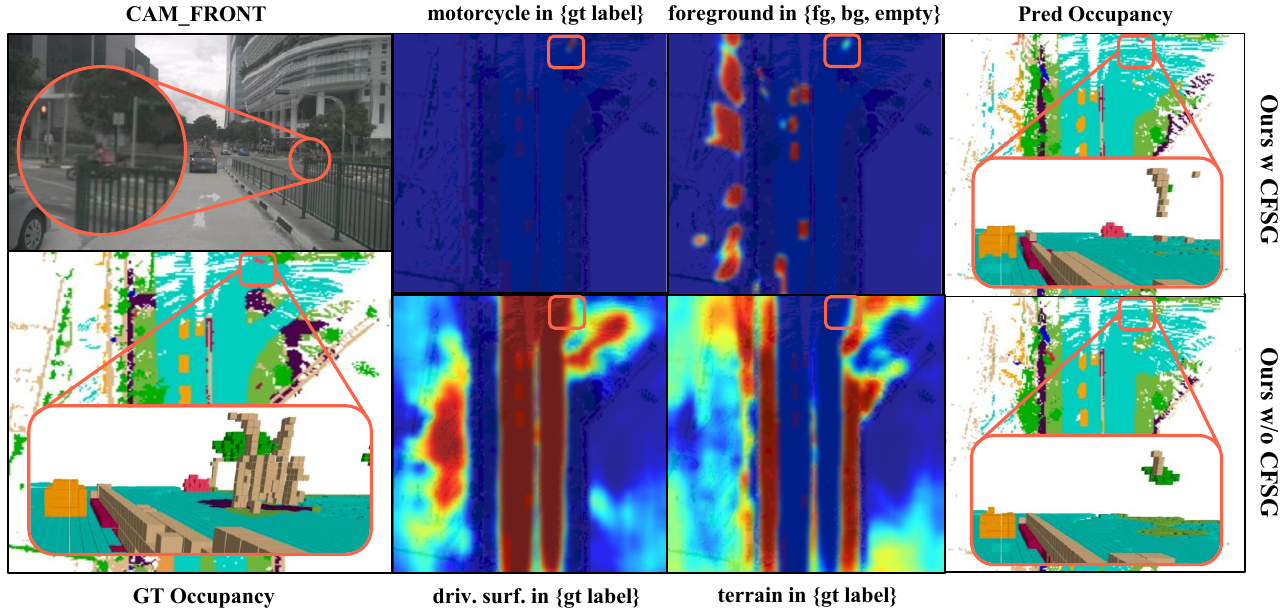}
    \setlength{\abovecaptionskip}{-1.0em}
    \caption{We visualize the heatmaps using different group masks. \textit{motorcycle in \{gt label\}} means the mask query is supervised by ground truth label, and \textit{foreground in \{fg, bg, empty\}} means the mask query is supervised by semantic group \textit{\{``foreground", ``background", ``empty"\}}. The rare category \textit{motorcycle} was successfully detected in various semantic groups, thereby enhancing the supervision signal.}
    \label{fig:group_heatmap}
    \vspace{-1.5em}
\end{figure*}

\label{subsec:ablation}
\textbf{The Effectiveness of Each Component.} The results are shown in Table~\ref{tab:ablation_compo}, we can observe that all components make their own performance contributions. The baseline achieves 70.36\% of IoU and 36.01\% of mIoU without long-term temporal information. We first integrated the Geometry-aware Occupancy Encoder (GOE) into the baseline model, which brings 0.53\% and 1.98\% performance gain in IoU and mIoU. By converting the 3D occupancy task to mask classification with the help of a Transformer decoder (TD), the network's semantic segmentation capability has been significantly enhanced. By using both GOE and TD, the network can excel in both geometric completion and semantic segmentation, outperforming the baseline by 1.38\% of IoU and 4.21\% of mIou. Moreover, the Coarse-to-Fine Semantic Grouping further enhances rare object recognition and achieves 41.05\% mIoU scores while retaining essentially the same geometry completion capability.

\textbf{The Effectiveness of Compact Occupancy Representation.} To further demonstrate the effect of a compact occupancy representation, we conducted an experiment where we used different representations of the occupancy features. As shown in Fig.~\ref{fig:proxy_exp}~(a), by introducing a U-net to bridge the explicit and implicit view transformation in a voxel representation, we have achieved a balance between performance and computational efficiency. Qualitative results illustrated in Fig.~\ref{fig:EI_ablation} showcase that the compact occupancy representation is able to bring improvements in geometric completion, especially for slender objects such as pedestrians and poles. Moreover, the compact occupancy representation exhibits robustness to occlusions.

\textbf{The Effectiveness of Semantic-aware Group Decoder.} In Fig.~\ref{fig:god_longtail_cls}, we compare the results of adopting the Semantic-aware Group Decoder (SGD) according to the label distribution. It is clear to see that there exists a significant class imbalance phenomenon in the dataset, for example, where the 6 background categories account for 93.8\% of the total labels. SGD significantly enhances the semantic discriminability of the compact occupancy representation through the transformer decoder and balancing supervision within each group using coarse-to-fine semantic grouping.

\textbf{The Effectiveness of Coarse-to-Fine Semantic Grouping.} To further demonstrate the impact of CFSG, we compare the performance impact of using different numbers of semantic groups. As shown in Table~\ref{tab:ablation_group_num}, replicating the original ground truth labels ten times like~\cite{chen2023group} to formulate a one-to-many assignment and increase the number of model parameters does not result in performance improvement. However, adding a simple \textit{\{``foreground", ``background", ``empty"\}} group can give a significant performance boost. Moreover, we generated heatmaps to visualize query masks in various groups. As illustrated in Fig.~\ref{fig:group_heatmap}, the rare class motorcycle is correctly detected in both the group of \textit{\{gt~label\}} or \textit{\{``foreground", ``background", ``empty"\}}, which strengthens the semantic supervision of this class. In contrast, the network without using CFSG experienced suppression of its response values by the background class at that position.
\section{Conclusion}
\label{sec:conclusion}
In this paper, we have presented COTR, a Compact Occupancy TRansformer for vision-based 3D occupancy prediction. For a holistic understanding of the 3D scene, we construct a compact geometry- and semantic-aware 3D occupancy representation through efficient explicit-implicit view transformation and coarse-to-fine semantic grouping. We evaluate COTR with several prevailing baselines and achieve state-of-the-art performance on nuScenes. We hope COTR can motivate further research in vision-based 3D occupancy prediction and its applications in autonomous vehicle perception.

\noindent
\textbf{Acknowledgements} This work is supported by the National Natural Science Foundation of China No.62302167, U23A20343, 62222602, 62206293, 62176224, 62106075, and 62006139, Shanghai Sailing Program (23YF1410500), Natural Science Foundation of Shanghai (23ZR1420400), Science and Technology Commission of Shanghai No.21511100700, Natural Science Foundation of Chongqing, China (CSTB2023NSCQ-JQX0007, CSTB2023NSCQ-MSX0137), CCF-Tencent Rhino-Bird Young Faculty Open Research Fund (RAGR20230121) and CAAI-Huawei MindSpore Open Fund.
{
    \small
    \bibliographystyle{ieeenat_fullname}
    \bibliography{main}
}
\clearpage
\setcounter{page}{1}

\twocolumn[{
\renewcommand\twocolumn[1][]{#1}
\maketitlesupplementary
\definecolor{nbarrier}{RGB}{112, 128, 144}
\definecolor{nbicycle}{RGB}{220, 20, 60}
\definecolor{nbus}{RGB}{255, 127, 80}
\definecolor{ncar}{RGB}{255, 158, 0}
\definecolor{nconstruct}{RGB}{233, 150, 70}
\definecolor{nmotor}{RGB}{255, 61, 99}
\definecolor{npedestrian}{RGB}{0, 0, 230}
\definecolor{ntraffic}{RGB}{47, 79, 79}
\definecolor{ntrailer}{RGB}{255, 140, 0}
\definecolor{ntruck}{RGB}{255, 99, 71}
\definecolor{ndriveable}{RGB}{0, 207, 191}
\definecolor{nother}{RGB}{175, 0, 75}
\definecolor{nsidewalk}{RGB}{75, 0, 75}
\definecolor{nterrain}{RGB}{112, 180, 60}
\definecolor{nmanmade}{RGB}{222, 184, 135}
\definecolor{nvegetation}{RGB}{0, 175, 0}
\definecolor{nothers}{RGB}{0, 0, 0}

\begin{table}[H]
	\footnotesize
    \begin{center}
    \setlength{\tabcolsep}{0.0057\linewidth}
	\begin{tabular}{l|c|c|cc| c c c c c c c c c c c c c c c c c}
		\toprule
		Method
		& Epoch & Mask & IoU & mIoU
        & \rotatebox{90}{\textcolor{nothers}{$\blacksquare$} others}
        
		& \rotatebox{90}{\textcolor{nbarrier}{$\blacksquare$} barrier}
		
		& \rotatebox{90}{\textcolor{nbicycle}{$\blacksquare$} bicycle}
		
		& \rotatebox{90}{\textcolor{nbus}{$\blacksquare$} bus}

		& \rotatebox{90}{\textcolor{ncar}{$\blacksquare$} car}

		& \rotatebox{90}{\textcolor{nconstruct}{$\blacksquare$} const. veh.}

		& \rotatebox{90}{\textcolor{nmotor}{$\blacksquare$} motorcycle}

		& \rotatebox{90}{\textcolor{npedestrian}{$\blacksquare$} pedestrian}

		& \rotatebox{90}{\textcolor{ntraffic}{$\blacksquare$} traffic cone}

		& \rotatebox{90}{\textcolor{ntrailer}{$\blacksquare$} trailer}

		& \rotatebox{90}{\textcolor{ntruck}{$\blacksquare$} truck}

		& \rotatebox{90}{\textcolor{ndriveable}{$\blacksquare$} drive. suf.}

		& \rotatebox{90}{\textcolor{nother}{$\blacksquare$} other flat}

		& \rotatebox{90}{\textcolor{nsidewalk}{$\blacksquare$} sidewalk}

		& \rotatebox{90}{\textcolor{nterrain}{$\blacksquare$} terrain}

		& \rotatebox{90}{\textcolor{nmanmade}{$\blacksquare$} manmade}

		& \rotatebox{90}{\textcolor{nvegetation}{$\blacksquare$} vegetation}

		\\
		\midrule
        MonoScene~\cite{cao2022monoscene}& 24 & \ding{55} & - & 6.1 & 1.75 & 7.23 & 4.26 & 4.93 & 9.38 & 5.67 & 3.98 & 3.01 & 5.90 & 4.45 & 7.17 & 14.91 & 6.32 & 7.92 & 7.43 & 1.01 & 7.65 \\
        OccFormer\cite{Zhang_2023_ICCV} & 24 & \ding{55} & 30.1 & 20.4 & 6.62 & 32.57 & 13.13 & 20.37 & 37.12 & 5.04 & 14.02 & 21.01 & 16.96 & 9.34 & 20.64 & 40.89 & 27.02 & 27.43 & 18.65 & 18.78 & 16.90 \\
        BEVFormer~\cite{li2022bevformer} & 24 & \ding{55} & - & 26.9 & 5.85 & 37.83 & 17.87 & 40.44 & 42.43 & 7.36 & 23.88 & 21.81 & 20.98 & 22.38 & 30.70 & 55.35 & 28.36 & 36.0 & 28.06 & 20.04 & 17.69 \\
        CTF-Occ~\cite{tian2023occ3d} & 24 & \ding{55} & - & 28.5 & 8.09 & 39.33 & 20.56 & 38.29 & 42.24 & 16.93 & 24.52 & 22.72 & 21.05 & 22.98 & 31.11 & 53.33 & 33.84 & 37.98 & 33.23 & 20.79 & 18.0 \\
        VoxFormer~\cite{li2023voxformer} & 24 & \ding{52} & - & 40.7 & - & - & - & - & - & - & - & - & - & - & - & - & - & - & - & - & - \\
        SurroundOcc~\cite{wei2023surroundocc} & 24 & \ding{52} & - & 40.7 & - & - & - & - & - & - & - & - & - & - & - & - & - & - & - & - & - \\
        FBOcc~\cite{li2023fb} & 20 & \ding{52} & - & 42.1 & 14.30 & 49.71 & 30.0 & 46.62 & 51.54 & 29.3 & 29.13 & 29.35 & 30.48 & 34.97 & 39.36 & 83.07 & 47.16 & 55.62 & 59.88 & 44.89 & 39.58 \\ 
        \midrule
        TPVFormer~\cite{huang2023tri} & 24 & \ding{52} & 66.8 & 34.2 & 7.68 & 44.01 & 17.66 & 40.88 & 46.98 & 15.06 & 20.54 & 24.69 & 24.66 & 24.26 & 29.28 & 79.27 & 40.65 & 48.49 & 49.44 & 32.63 & 29.82 \\
        \rowcolor[HTML]{EFEFEF}
        + COTR (Res-50) & 24 & \ding{52} & \textbf{70.6} & \textbf{39.3} & \textbf{11.66} & \textbf{45.47} & \textbf{25.34} & \textbf{41.71} & \textbf{50.77} & \textbf{27.39} & \textbf{26.30} & \textbf{27.76} & \textbf{29.71} & \textbf{33.04} & \textbf{37.76} & \textbf{80.52} & \textbf{41.67} & \textbf{50.82} & \textbf{54.54} & \textbf{44.91} & \textbf{38.27} \\
        SurroundOcc~\cite{wei2023surroundocc} & 24 & \ding{52} & 65.5 & 34.6 & 9.51 & 38.50 & 22.08 & 39.82 & 47.04 & 20.45 & 22.48 & 23.78 & 23.00 & 27.29 & 34.27 & 78.32 & 36.99 & 46.27 & 49.71 & 35.93 & 32.06 \\
        \rowcolor[HTML]{EFEFEF}
         + COTR (Res-50) & 24 & \ding{52} & \textbf{71.0} & \textbf{39.3} & \textbf{11.37} & \textbf{45.90} & \textbf{25.97} & \textbf{43.08} & \textbf{51.56} & \textbf{25.55} & \textbf{26.21} & \textbf{26.53} & \textbf{29.48} & \textbf{33.65} & \textbf{38.87} & \textbf{80.45} & \textbf{40.34} & \textbf{50.86} & \textbf{53.88} & \textbf{45.37} & \textbf{38.94}\\
        \midrule
        OccFormer\cite{Zhang_2023_ICCV} & 24 & \ding{52} & 70.1 & 37.4 & 9.15 & 45.84 & 18.20 & 42.80 & 50.27 & 24.00 & 20.80 & 22.86 & 20.98 & 31.94 & 38.13 & 80.13 & 38.24 & 50.83 & 54.3 & 46.41 & 40.15 \\
        \rowcolor[HTML]{EFEFEF}
        + COTR (Res-50) & 24 & \ding{52} & \textbf{71.7} & \textbf{41.2} & \textbf{12.19} & \textbf{48.47} & \textbf{27.81} & \textbf{44.28} & \textbf{52.82} & \textbf{28.70} & \textbf{28.16} & \textbf{28.95} & \textbf{31.32} & \textbf{35.01} & \textbf{39.93} & \textbf{81.54} & \textbf{42.05} & \textbf{53.44} & \textbf{56.22} & \textbf{47.37} & \textbf{41.38} \\
        \midrule
        OccFormer\cite{Zhang_2023_ICCV} & 24 & \ding{52} & 70.1 & 37.4 & 9.15 & 45.84 & 18.20 & 42.80 & 50.27 & 24.00 & 20.80 & 22.86 & 20.98 & 31.94 & 38.13 & 80.13 & 38.24 & 50.83 & 54.3 & 46.41 & 40.15 \\
        \rowcolor[HTML]{EFEFEF}
        + COTR (Res-50) & 24 & \ding{52} & \textbf{71.7} & \textbf{41.2} & \textbf{12.19} & \textbf{48.47} & \textbf{27.81} & \textbf{44.28} & \textbf{52.82} & \textbf{28.70} & \textbf{28.16} & \textbf{28.95} & \textbf{31.32} & \textbf{35.01} & \textbf{39.93} & \textbf{81.54} & \textbf{42.05} & \textbf{53.44} & \textbf{56.22} & \textbf{47.37} & \textbf{41.38} \\
        \midrule
        BEVDet4D~\cite{huang2022bevdet4d} & 24 & \ding{52} & 73.8 & 39.3 & 9.33 & 47.05 & 19.23 & 41.47 & 52.21 & 27.19 & 21.23 & 23.32 & 21.58 & 35.77 & 38.94 & 82.48 & 40.42 & 53.75 & 57.71 & 49.94 & 45.76 \\
        \rowcolor[HTML]{EFEFEF}
        + COTR (Res-50) & 24 & \ding{52} & \textbf{75.0} & \textbf{44.5} & \textbf{13.29} & \textbf{52.11} & \textbf{31.95} & \textbf{46.03} & \textbf{55.63} & \textbf{32.57} & \textbf{32.78} & \textbf{30.35} & \textbf{34.09} & \textbf{37.72} & \textbf{41.84} & \textbf{84.48} & \textbf{46.19} & \textbf{57.55} & \textbf{60.67} & \textbf{51.99} & \textbf{46.33} \\
        \midrule
        BEVDet4D~\cite{huang2022bevdet4d} & 36 & \ding{52} & 72.3 & 42.5 & 12.37 & 50.15 & 26.97 & \textbf{51.86} & 54.65 & 28.38 & 28.96 & 29.02 & 28.28 & 37.05 & 42.52 & 82.55 & 43.15 & 54.87 & 58.33 & 48.78 & 43.79 \\
        \rowcolor[HTML]{EFEFEF}
        + COTR (Swin-B) & 24 & \ding{52} & \textbf{74.9} & \textbf{46.2} & \textbf{14.85} & \textbf{53.25} & \textbf{35.19} & 50.83 & \textbf{57.25} & \textbf{35.36} & \textbf{34.06} & \textbf{33.54} & \textbf{37.14} & \textbf{38.99} & \textbf{44.97} & \textbf{84.46} & \textbf{48.73} & \textbf{57.60} & \textbf{61.08} & \textbf{51.61} & \textbf{46.72} \\
		\bottomrule
	\end{tabular}
    \end{center}
    \vspace{-10pt}
\end{table}

\captionsetup{belowskip=8pt}
\captionsetup{aboveskip=0pt}
\captionof{table}{\textbf{3D Occupancy prediction performance on the Occ3D-nuScenes dataset.} We present the IoU (geometry) and mean IoU (semantic) over categories and the IoUs (semantic) for different classes.}
\label{tab:appendix_nuscene_sota}
}]

\appendix

\section{Further Implementation Details}
\label{sec:further_implement}
In this section, we further elaborate on the implementation details of our COTR. 

\textbf{Geometry-aware Occupancy Encoder.} As we mentioned in Sec.~\ref{subsec:geo_occ_encoder}, the Explicit View Transformation generates a occupancy feature $O_{\rm E} \in \mathbb{R}^{32 \times 200 \times 200 \times 16}$. Then, we use a 3D-ResNet according to ~\cite{huang2022bevdet4d} to generate multi-scale occupancy features $O_{\rm E}^0 \in \mathbb{R}^{32 \times 200 \times 200 \times 16}, O_{\rm E}^1 \in \mathbb{R}^{64 \times 100 \times 100 \times 8}, O_{\rm E}^2 \in \mathbb{R}^{128 \times 50 \times 50 \times 4}$. Next, employ trilinear interpolation to sample multi-scale OCC features into the same size of $50 \times 50 \times 16$, followed by a concatenation and convolutional layer to construct the compact OCC representation $O_c \in \mathbb{R}^{256 \times 50 \times 50 \times 16}$. Finally, the compact OCC representation is fed into Implicit View-Transformation for further update. Since the compact OCC feature $O_c$ has been already initialized by EVT, we only use 1 Transformer layer in IVT. With the final prediction resolution being $200 \times 200 \times 16$, we use deconvolution layers to upsample the compact OCC feature to $O \in \mathbb{R}^{256 \times 200 \times 200 \times 16}$, which was only utilized for mask prediction in Semantic-aware Group Decoder. In order to counteract the loss of geometric details throughout the process of downsampling, we construct a U-net\cite{ronneberger2015u} architecture by concatenated multi-scale features $\{O_{\rm E}^i\}_{i=0}^3$ to the upsampled features.

\textbf{Loss Function.} During training, we used a total of 4 different loss functions:
\begin{equation}
    \mathcal{L} = \lambda_{{\rm depth}}\mathcal{L_{\rm depth}} + \lambda_{{\rm seg}}\mathcal{L_{\rm seg}} + \lambda_{{\rm mask\text{-}cls}}\mathcal{L_{\rm mask\text{-}cls}},
\end{equation}
where $\mathcal{L_{\rm depth}}$ is the depth estimation loss in the Image Featurizers following BEVDepth~\cite{li2023bevdepth}, $\mathcal{L_{\rm seg}}$ is a simple cross-entropy segmentation loss between a coarse prediction from $O$ and the ground truth label, and the $\lambda_{{\rm mask\text{-}cls}}$ loss combines a cross entropy classification loss and a binary mask loss for each predicted mask segment following MaskFormer~\cite{cheng2021per}. We set the hyper-parameters to $\lambda_{{\rm depth}}=\lambda_{{\rm mask\text{-}cls}}=1$ and $\lambda_{{\rm seg}}=10$.

\section{Further Experiments}
\label{sec:further_implement}
\textbf{Per-class comparison with SOTA.} We report more quantitative details in Table ~\ref{tab:appendix_nuscene_sota} about our experimental results for better comparison with other competitors. Besides TPVFormer~\cite{huang2023tri} and BEVDet4D~\cite{huang2022bevdet4d}, we also integrate COTR into SurroundOcc~\cite{wei2023surroundocc} and OccFormer~\cite{Zhang_2023_ICCV}. COTR yields significant performance improvements in both geometric completion and semantic segmentation, surpassing SurroundOcc and OccFormer by 5.5\%, 1.6\% in IoU and 4.7\%, 3.8\% in mIoU. Notably, a conspicuous amelioration primarily resides within small objects and rare objects, demonstrating that our approach can indeed apprehend finer geometric details, and substantially enhance semantic discernibility.

\begin{table}[t]
\resizebox{\linewidth}{!}{
\begin{tabular}{c|c|ccc}
\toprule
Rep. & Resolution & IoU (\%)   & mIoU (\%) & FLOPs (G)\\
\midrule
BEV & $200 \times 200$  & 70.48 & 37.51 & 65.74 \\
TPV & $200 \times (200, 16 \times 2)$ & 70.41 & 37.21 & 291.62\\
\midrule
\multirow{5}{*}{Voxel} & $200 \times 200 \times 16$ & - & - & 402.98\\
 & $100 \times 100 \times 8$ & 70.83 & 37.36 & 101.73\\
 & $50 \times 50 \times 4$ & 70.87 & 37.61 & \textbf{61.27} \\
 & $50 \times 50 \times 8$ & 70.71 & 37.45 & 67.11 \\
\rowcolor[HTML]{EFEFEF}
 & $50 \times 50 \times 16$ & \textbf{70.89} & \textbf{37.97} & 78.47\\
\bottomrule
\end{tabular}
}
\setlength{\abovecaptionskip}{5pt}
\caption{\textbf{Ablation study for different occupancy representation resolution.} All models are trained without Semantic-aware Group Decoder and long-term temporal information. We report the FLOPs of the Implicit View Transformation module.}
\label{tab:ablation_occrep}
\end{table}
\begin{table}[t]
\begin{center}
\setlength{\tabcolsep}{0.02\linewidth}
\begin{tabular}{ccc|cccc}
\toprule
\multicolumn{3}{c|}{Component} & \multicolumn{4}{c}{Computational Cost} \\
\midrule
GOE & TD & CFSG & \multicolumn{2}{c}{Params (M)} & \multicolumn{2}{c}{FLOPs (G)} \\
\midrule
 &  &  & 34.97 & - & 541.21 & -\\
\ding{52} &  &  & 35.84 & {\color{blue}+0.87} & 573.82 & {\color{blue}+32.61}\\
& \ding{52} &  & 36.21 & {\color{blue}+1.24} & 628.75 & {\color{blue}+87.54}\\
& \ding{52} & \ding{52} & 36.21 & {\color{blue}+1.24} & 628.75 & {\color{blue}+87.54}\\
\bottomrule
\end{tabular}
\end{center}
\setlength{\abovecaptionskip}{0pt}
\caption{\textbf{Ablation study on each component's computational cost.} All models are trained without long-term temporal information.}
\label{tab:ablation_compo_cost}
\end{table}

\textbf{Ablation study for different OCC resolution.} Table ~\ref{tab:ablation_occrep} compared different resolutions for OCC representations in our experiments. It is abundantly clear that the high-resolution ($200 \times 200 \times 16$) OCC representation incurs a massive computational overhead, with the FLOPs approximately $5\times$ that of the compact ($50 \times 50 \times 16$) OCC representation, respectively. Additionally, it appears that preserving the height information is beneficial for the task of occupancy prediction. Overall, our compact OCC representation strikes a balance between performance and computational overhead.

\textbf{Ablation study on computational cost.} As shown in Table ~\ref{tab:ablation_compo_cost}, our proposed COTR is an efficient approach in which each component does not add a significant amount of computational costs. It's worth noting that, since we only used the Coarse-to-Fine Semantic Grouping (CFSG) strategy during training and kept only one group during inference, CFSG doesn't introduce any extra overhead. More metrics are reported in Table~\ref{tab:efficient}. All baselines are concurrently equipped with EVT and IVT for a fare comparison.
\begin{table}[t]
\small
\setlength{\tabcolsep}{0.015\linewidth}
\begin{center}
\begin{tabular}{l|cccc}
\hline
method                  & Params (M) & FLOPs (G) & LT & Latency (s) \\ \hline
TPVFormer        & 54.05      & 972.75    &  \ding{55}  & 0.59 \\
\rowcolor[HTML]{EFEFEF}
+ COTR (R50)  & 53.30       & 784.85    &  \ding{55}  &  0.43   \\ \hline
BEVDet4D $\dagger$         & 35.67      & 924.07    &  \ding{55}  & 0.67 \\
\rowcolor[HTML]{EFEFEF}
+ COTR (R50)  & 38.87      & 747.26    &  \ding{55}  & 0.42 \\ \hline
BEVDet4D           & 35.67      & 1049.52    &  \ding{52}  & 1.89 \\
\rowcolor[HTML]{EFEFEF}
+ COTR (R50)  & 38.87      & 747.26    &  \ding{52}  & 1.43 \\ \hline
BEVDet4D         & 121.28     & 4195.12   &  \ding{55}  & 1.67 \\
\rowcolor[HTML]{EFEFEF}
+ COTR (SwinB)& 104.99     & 3761.06   &  \ding{55}  & 1.43 \\ \hline
\end{tabular}
\end{center}
\setlength{\belowcaptionskip}{0.0cm}
\setlength{\abovecaptionskip}{-0.2cm}
\caption{\textbf{Ablation study on efficiency.} All models are tested on a RTX A6000 GPU. $\dagger$ input size of $256 \times 704$, others same as Tab.~1. }
\vspace{-1.5em}
\label{tab:efficient}
\end{table}

\section{Visualization}
\label{sec:vis}
In this section, we provide more visualization results of our method.

\begin{figure*}[]
    \centering
    \includegraphics[width=\linewidth]{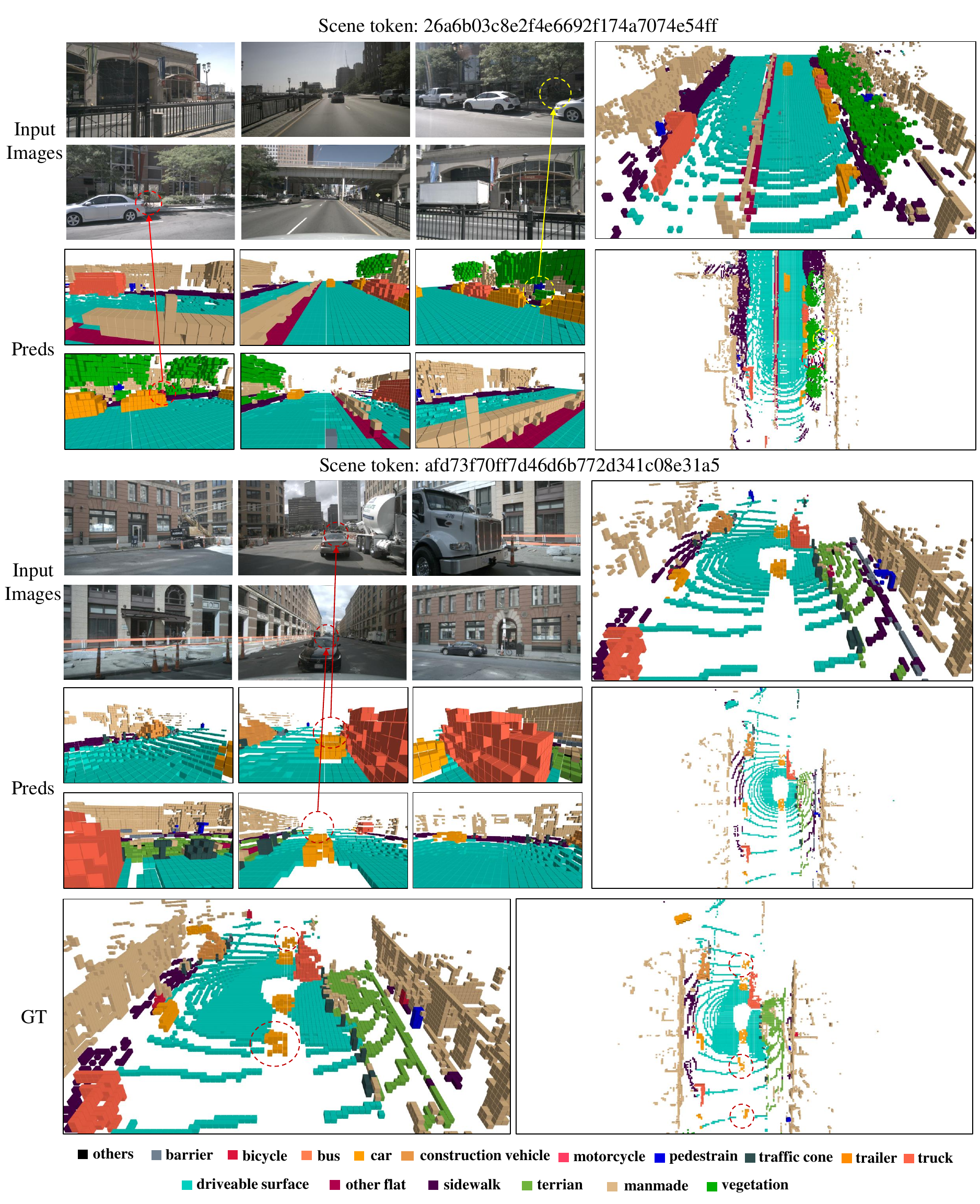}
    \caption{Visualizations for occlusion on OCC3D-nuScenes validation set. For each scene, the six images in the \textit{``Input Images''} line left are the inputs to our model captured by font-left, front, front-right, back-right, back, and back-right cameras. The six images in the \textit{``Preds''} line left denote our prediction results with the corresponding views as the inputs. The two images on the right provide a global view of our predictions. The two images in the \textit{``GT''} line provide a global view of ground truth.}
    \label{fig:vis_occlusion_failure}
\end{figure*}

\textbf{Visual Ablations on Occluded Scenes.} To validate the robustness of our method in handling occluded scenes, we present additional visual results. As shown in the first scene in Fig.~\ref{fig:vis_occlusion_failure}, our method without using long-term temporal information, successfully detects small objects (such as pedestrians and bicycles) within a limited occlusion range. However, in the second scene, when a significant portion of the vehicle is obscured, the model struggles to correctly identify occluded objects due to the constrained camera perspective.

\begin{figure*}[]
    \centering
    \includegraphics[width=\linewidth]{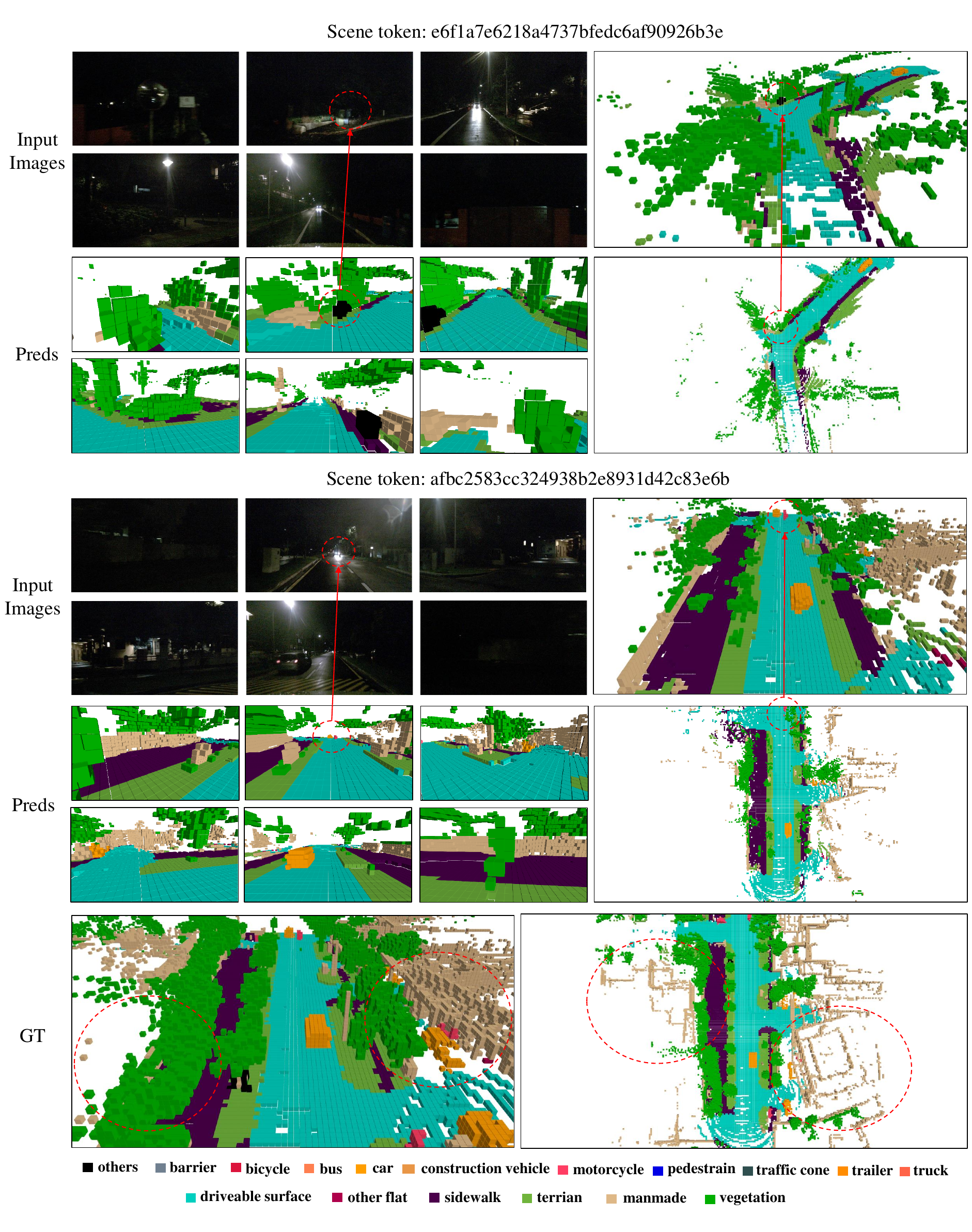}
    \caption{Visualizations for low-light environments on OCC3D-nuScenes validation set.}
    \label{fig:vis_night}
\end{figure*}

\textbf{Visual Ablations on Low-light Scenes.} To validate the robustness of our method in handling low-light environments, we present additional visual results. As illustrated in the first scenario of Fig.~\ref{fig:vis_night}, our model is capable of successfully detecting unknown objects in the dark. However, the second scenario shows that while our model can detect small objects in the dark from a distance, it fails to predict successfully when part of the camera is nearly completely obscured by darkness. This limitation is primarily due to the camera's perception capabilities and other modalities such as LiDAR or Radar might be required to aid successful detection.
\end{document}